# When the implementation matters more than the algorithm: NSGA-II and other multi-objective optimizers in engineering design tools

**Luciano Ambrosini**

LA Architecture & Computational Design Consulting, Naples, Italy

Corresponding author: Luciano Ambrosini, luciano.ambrosini@outlook.com. ORCID: 0000-0003-1529-2694

## Abstract

Engineering design tools increasingly embed optimizers as ready-made components that users select by algorithm name, while operator settings and bound, integer and duplicate handling stay fixed inside each tool. This study tests whether the name predicts the result. A tool-agnostic harness exposes one deterministic black box, an emulator of a published 22-variable mixed-integer building-performance problem, to the native optimizers of three Grasshopper plug-ins, to canonical NSGA-II, SMS-EMOA and GDE3 from Python libraries, and to the model-based RBFMOpt. Eighteen configurations were run at 5,000 evaluations under three population allocations, with statistics fixed in advance. Four NSGA-II implementations differed by up to 5% in hypervolume, more than three canonical algorithms of different families differed from one another, and their response to population size differed by up to 10 percentage points; for two MOEA/D implementations, by 28. On ZDT test functions the same NSGA-II implementations differed up to 470-fold in IGD. An ablation in one open-source NSGA-II reproduced the behaviour of one tool from its documented settings but not the gaps of the other two, so documented parameters do not identify an implementation. Counting the budget in wall-clock time reversed the ranking of the model-based optimizer, and reporting the final population instead of the archive understated some configurations by up to a third; most differences fell below the surrogate's error. For optimizers embedded in design-support systems, the implementation is part of the method; a reporting checklist is derived: implementation and version, effective settings, budget in time and evaluations, and archive-based indicators.

## 1. Introduction

Optimization reaches most designers as a component of a design environment rather than as a library. Such environments act as design-support systems in which the optimizer is one embedded component among geometry, simulation and visualization, and the user interacts with it through a small interface rather than through code. In parametric tools such as Grasshopper, the user drags an optimizer onto the canvas, connects the design variables and the objectives, and chooses an algorithm from a list of names: NSGA-II, MOEA/D, a differential-evolution or a model-based method. The operator settings, the handling of bounds and integer variables and the treatment of duplicate solutions are fixed by the plug-in and are rarely visible in its interface. Studies that use these tools typically report the algorithm name, keep the defaults and present the final population of a single run [1, 2]. This practice assumes that the name of an algorithm identifies its behaviour.

The evolutionary computation community has shown that this assumption is fragile. Results depend on implementation details that publications often omit [3, 4], and properties of an implementation can bias benchmark conclusions without the algorithm itself being at fault [5, 6]. The handling of out-of-bound solutions alone changes the behaviour of differential evolution and particle swarm optimization [7–9], and implementations of NSGA-II, GDE3 and MOEA/D in different optimization frameworks give significantly different results under nominally identical settings [10], as do single-objective algorithms [11]. These studies concern libraries used by optimization researchers on analytic test functions, where the settings can be inspected and changed. Engineering design software differs in three respects: the user cannot inspect or change most settings, different tools offer the same algorithm with different defaults, and what matters is the consequence in the units of the design problem. Whether implementation then matters as much as the choice of algorithm, whether the documented settings explain the differences, and whether they change a design has not been measured in a controlled way. What is missing is a test of optimizers as they are deployed in design systems, rather than as they are published.

Building performance optimization is a demanding test case. Energy demand, thermal comfort and daylight conflict, so the designer needs a set of trade-offs, and population-based metaheuristics, above all NSGA-II [12], are the default in research and practice [13, 14]. Benchmarks on building problems have compared global search algorithms [15, 16] and model-based against population-based methods at small budgets [17–19]; like most benchmarks, they compare algorithms by name. Machine-learning surrogates now replace the building simulation in many studies [20], and evaluations that cost milliseconds change the computational picture: thousands of evaluations are affordable, repeated runs fit in a working session, and the overhead of the optimizer can dominate the run time [17].

Zorn et al. [2] published a problem that makes a controlled comparison possible: a 22-variable, mixed-integer shoebox office with total energy demand, overheating degree-hours and daylight factor as objectives, a surrogate version and the full data set [21]. This paper uses an emulator of that surrogate as one deterministic black box, queried identically by every optimizer, so that any difference in outcome comes from the optimizer and its implementation and not from the evaluation. Four questions are asked:

- **RQ1.** Does the algorithm or its implementation decide the result? NSGA-II is run in four independent implementations, MOEA/D and GDE3 in two.
- **RQ2.** Does advice on the population allocation transfer between implementations of the same algorithm?
- **RQ3.** How do the rankings change when the budget is counted in wall-clock time instead of evaluations?

- **RQ4.** How much do the reported set and the indicator change the conclusions, and are the differences large enough to change a design?

The contributions are: (i) a tool-agnostic evaluation harness in which the native optimizers of three Grasshopper plug-ins and three canonical library algorithms query the same emulator, with C# and NumPy inferencers that agree to $10^{-12}$ (Section 2); (ii) a controlled comparison of eighteen configurations at three population allocations under a statistical protocol fixed before the final campaign, together with a side-by-side account of the settings that differ between implementations of the same algorithm, a check of the implementation contrasts on analytic test functions, and an ablation that tests whether the documented settings explain the differences (Sections 3–6); (iii) an analysis of the differences in design terms, against the error of the surrogate, and of whether rankings on analytic test functions transfer to the building problem (Sections 5.4–5.7); and (iv) a reporting checklist for optimizers embedded in design software, derived from these results (Section 6.4). In line with the no-free-lunch argument [22] and benchmarking guidance [4, 23], the conclusions on algorithm rankings are specific to this problem class; the findings on implementation, budget and reporting concern the software and transfer to any tool that exposes optimizers by name. All evaluations, the emulator and the analysis code are released, and every configuration except those of the not-yet-released SilverHawk plug-in can be re-run.

## 2. Benchmark problem and evaluation harness

### *2.1. Problem*

The problem is the one defined by Zorn et al. [2] with engineers from Transsolar: a south-facing, 27 m² single-zone office in Stuttgart with a window of variable sill height and a static shading device of five rows of triangulated elements. Three objectives are optimized: total annual energy demand $Q_{tot}$ [kWh/(m²·a)] and overheating degree-hours $K_h$ [K·h/a], both minimized, and daylight factor DF [–], maximized; the definitions and simulation settings are those of the original study [2]. For hypervolume computation the objectives are expressed in minimization form, ($Q_{tot}$, $K_h$, −DF).

The 22 decision variables (Fig. 1; Supplementary Table S1) are 18 continuous ones (insulation thickness, night-time air-change rate, window sill height, and the X, Y, Z coordinates of the five shading rows) and four binary construction and system choices, so the problem is mixed-integer. For the Z coordinates, Table 1 of [2] states [0, 1], but every released surrogate-problem sample and every TRNSYS evaluation lies in [0, 0.5] [21]; [0, 0.5] is adopted here. For insulation thickness the declared bound of 0.49 m is kept, although the released samples stop at 0.44 m; Section 5.5 checks this range.

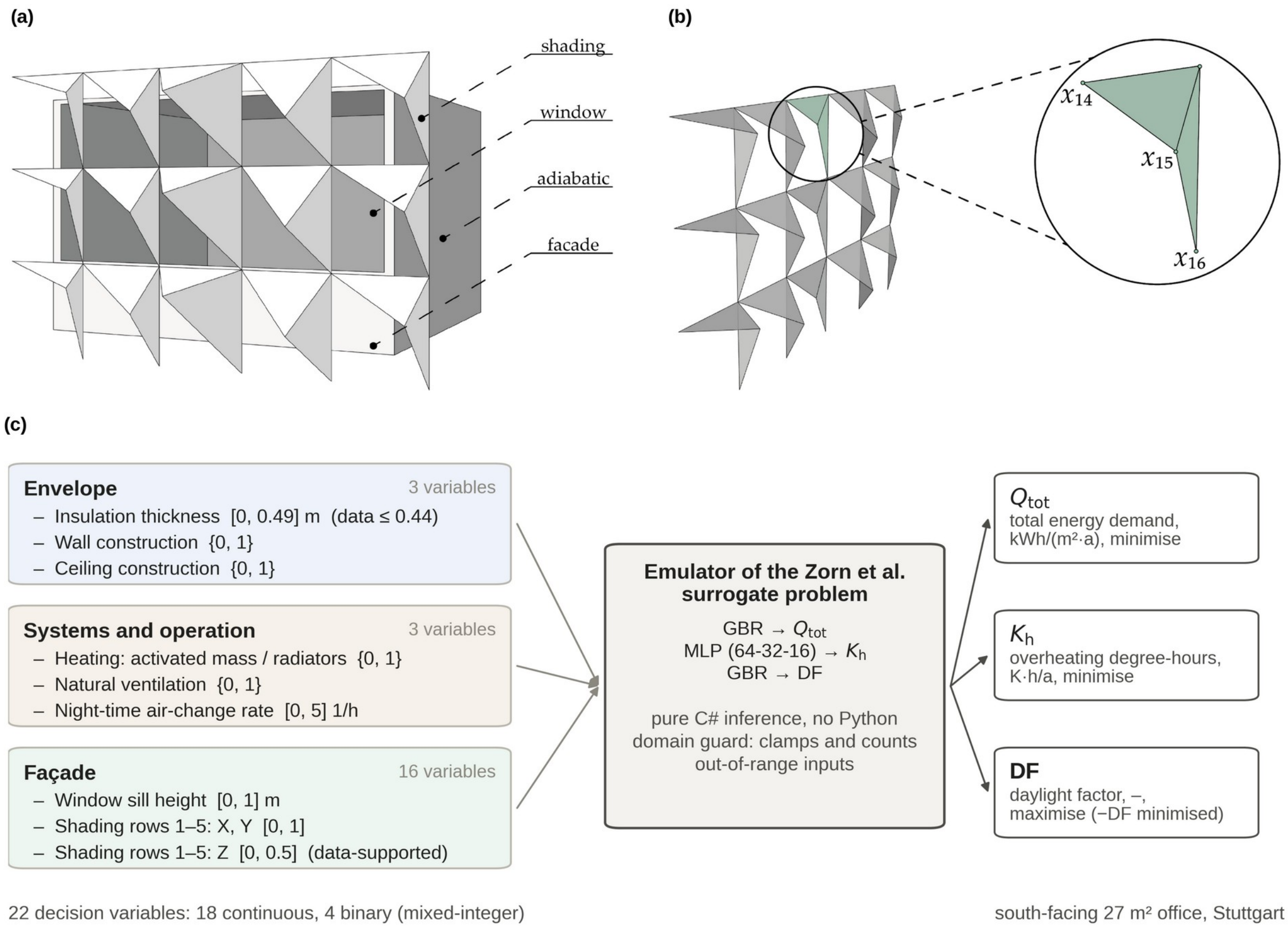

**Fig. 1.** Benchmark problem. (a) The south-facing shoebox office with its static shading device; (b) shading geometry; the highlighted points are the X, Y, Z coordinates of row 3 ($x_{14}$–$x_{16}$ in the notation of [2]); (c) the 22 decision variables, grouped by building component, mapped by the emulator of the surrogate problem to the three objectives, with the bounds enforced in this study. (a, b) reproduced from Zorn et al. [2] under CC BY 4.0; (c) this study.

### *2.2. Emulator of the surrogate problem and its scope*

Zorn et al. report a TRNSYS benchmark and a surrogate benchmark in which gradient-boosting regressors (GBR) predict $Q_{tot}$ and DF and a neural network predicts $K_h$ [2]. The released set of 10,001 Latin hypercube samples (SuMo_Random) is labelled by these surrogate models, not by TRNSYS: the samples were logged within 17 min, against about 7 min per TRNSYS evaluation, and their DF and $K_h$ never exceed about 0.34 and 1,500 K·h/a, whereas the TRNSYS trajectories reach 0.56 and 15,000 K·h/a [21].

The models of this study are therefore **emulators of the published surrogate problem**, not a new BPS surrogate. They follow the model classes and preprocessing of Zorn et al.: interquartile-range outlier removal, standard scaling, a stratified 80/20 split on the binary variables, 5-fold cross-validation and their hyperparameter grid for the two GBR models (selected: 300 estimators, depth 3); $K_h$ is predicted by a multilayer perceptron (64-32-16 ReLU units). The fitted models are exported to JSON and evaluated by a C# inferencer inside Grasshopper and by a NumPy inferencer for the library algorithms; both reproduce the scikit-learn predictions to within $10^{-12}$ (Supplementary S1).

Table 1 reports the fidelity. On 200,019 points of the original surrogate-problem optimization runs (OOS-SuMo), which sample the regions optimizers actually visit, the emulator reproduces the surrogate problem closely ($R^2 \geq 0.96$). Against the TRNSYS results of the original runs (OOS-BPS), its accuracy for $Q_{tot}$ matches that of the original model, but for $K_h$ and DF it is poor, because the surrogate problem itself spans a much narrower response range than the physical one. This fixes the scope of the study. The emulator is a realistic 22-dimensional, mixed-integer response surface with the variable structure and objective conflicts of a building problem, and all optimizers see the same one; the retrieved designs are not recommendations for the physical office and would need re-simulation before use [2]. The question addressed is how choices of algorithm, implementation, budget and reporting change what a designer retrieves from such a surface, and the RMSE against TRNSYS (Table 1) serves in Section 5.4 as the yardstick for whether a difference between optimizers is large enough to matter.

**Table 1.** Emulator accuracy. CV = 5-fold cross-validation; OOS-SuMo = 200,019 points of the original surrogate-problem runs; OOS-BPS = 17,309 TRNSYS evaluations of the original simulation-based runs. Zorn et al.: test-set results of their models [2, Table 4].

| **Target** | **Model** | **CV $R^2$** | **OOS-SuMo $R^2$** | **OOS-BPS $R^2$** | **OOS-BPS RMSE** | **Zorn et al. $R^2$ / RMSE** |
|---|---|---|---|---|---|---|
| $Q_{tot}$ [kWh/(m²·a)] | GBR | 0.990 | 0.962 | 0.746 | 2.84 | 0.73 / 3.14 |
| $K_h$ [K·h/a] | MLP | 0.996 | 0.990 | 0.144 | 851 | 0.92 / 81.2 |
| DF [–] | GBR | 0.981 | 0.977 | −0.152 | 0.130 | 0.78 / 0.033 |

### *2.3. A shared black box for all solvers*

In Grasshopper an optimizer sees a problem only through sliders that it sets and numeric outputs that it reads. A benchmark component sits between the 22 sliders and the three objective outputs of a single definition (Fig. 2). On set-up it writes the variable bounds into the sliders, so every optimizer reads the same domain; on each solution it passes the slider values through a domain guard, which clamps and counts out-of-range inputs, and through the emulator. A utility component sets the sliders to random positions before each run, so that tools that start from the current slider state do not begin from the same point. The optimizers of the three tools are wired to the same sliders and outputs and run in turn on the same definition, so the evaluation, the domain and the per-evaluation cost are identical; the remaining differences are the solvers and the way each tool drives the canvas.

The library algorithms run outside Grasshopper on the NumPy inferencer, with the same bounds and the same rounding of the binary variables. The algorithm code of one of the tools (SilverHawk) is also linked into a headless console, which runs the same code and emulator with seeded runs and logs every evaluation in order; it provides the anytime curves of Section 5.3 and a platform-consistency check of the canvas results.

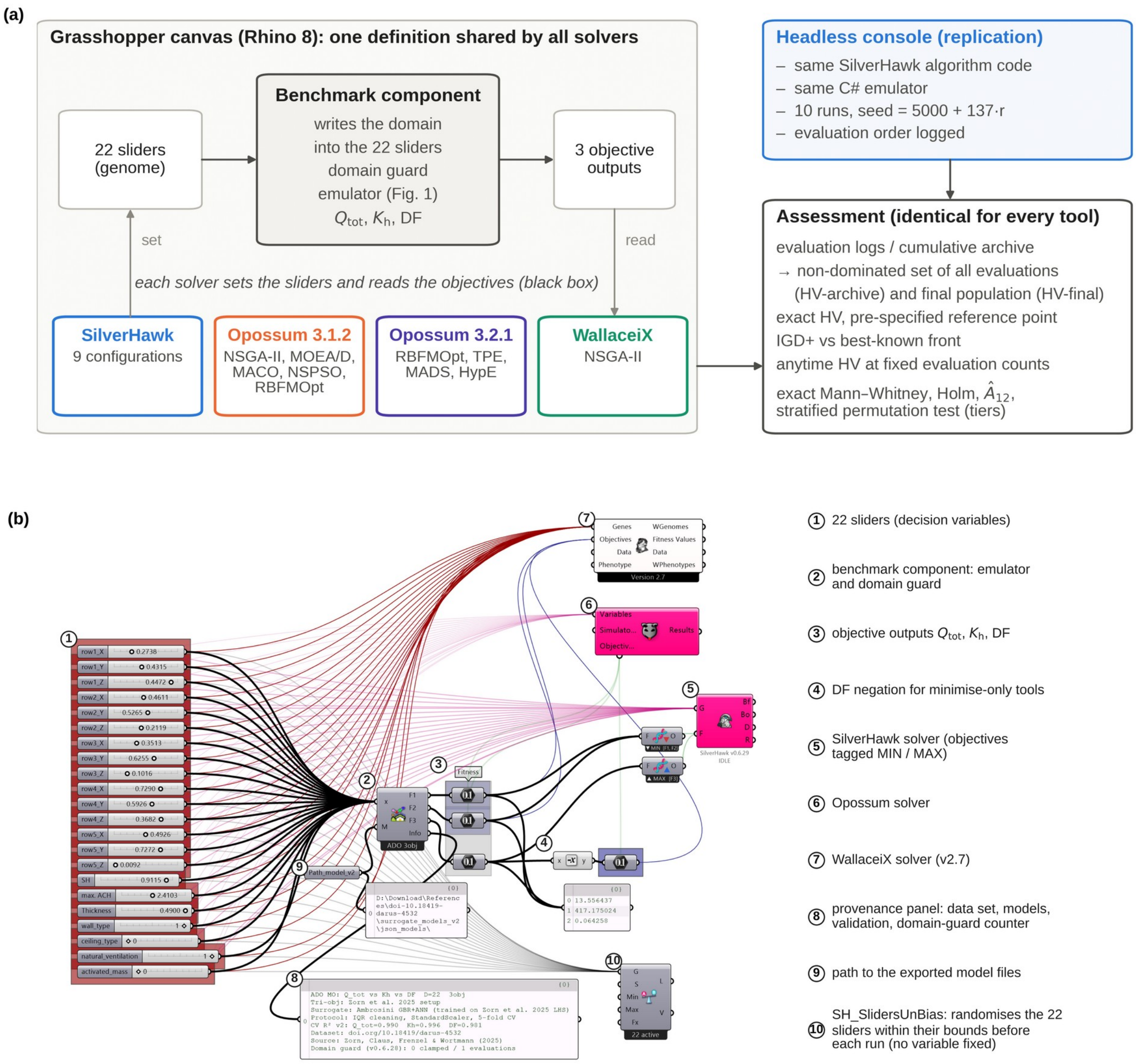

**Fig. 2.** Benchmark harness. (a) One Grasshopper definition is shared by the solvers of three tools: each tool sets the 22 sliders and reads the three objective outputs of the benchmark component, which holds the domain, the domain guard and the emulator; the headless console runs the same algorithm code and emulator, and all runs are assessed with the same procedure. The library algorithms evaluate the same emulator outside Grasshopper (Section 2.3). (b) The Grasshopper definition used in the campaign, with the numbered elements listed on the right.

## 3. Optimizers and experimental design

### *3.1. Configurations*

The roster combines three sources (Table 2). The first is the native optimizers of the Grasshopper tools that designers use for multi-objective problems: Opossum, whose metaheuristics come from the pagmo library [24], WallaceiX [25] and SilverHawk, a plug-in developed by the author (see the Declaration of competing interest). The second is canonical implementations of three algorithms from general-purpose Python libraries, run with their library defaults and without any modification: NSGA-II and SMS-EMOA from pymoo [12, 26, 27] and GDE3 from pymoode [28, 29]. They serve as an independent reference for the variation and selection mechanisms that the tool configurations build on, and they make it possible to separate the effect of an algorithm from that of its implementation. The third is the model-based RBFMOpt [30, 31], the method that led the original simulation-based benchmark. In the library runs the four binary variables are treated as continuous in [0, 1] and rounded when the emulator is evaluated, whereas the Grasshopper tools set them through integer sliders.

NSGA-II is thus present in four implementations, MOEA/D in two and GDE3 in two. The tool configurations are run in the release designers download: Opossum 3.1.2, the last release that ships the pagmo metaheuristics (release 3.2.0 replaced them with TPE, MADS and HypE, reported in Supplementary S3), WallaceiX 2.7 and SilverHawk 0.6.28–0.6.29.

**Table 2.** Configurations by algorithm family. Tool configurations run at the defaults of the release; library configurations run at the library defaults. † = literature algorithm with modifications documented in Supplementary S2; ★ = configuration developed in SilverHawk.

| **Family** | **Configuration (tool or library)** | **Key mechanism** | **Reference** |
|---|---|---|---|
| Genetic, non-dominated sorting | NSGA-II (pymoo; Opossum; WallaceiX; SilverHawk †) | SBX + polynomial mutation; non-dominated sorting + crowding | [12] |
| | NSGA-III (SilverHawk †) | Das–Dennis reference directions | [32] |
| Genetic, hypervolume selection | SMS-EMOA (pymoo) | Last front truncated by exclusive hypervolume contribution | [27] |
| Differential evolution (DE) | GDE3 (pymoode) | DE/rand/1/bin; pairwise dominance selection; crowding | [29] |
| | GDE3 (SilverHawk †) | as above + jDE self-adaptation [33], Solis–Wets refinement [34], archive mating | [29] |
| | OSPREY (SilverHawk ★) | GDE3 † with hypervolume-contribution truncation, as in SMS-EMOA | [27] |
| | EAGLE (SilverHawk ★) | jDE/rand/1/bin + Lévy-flight swarm trials + Solis–Wets; hypervolume-contribution truncation | – |
| Particle swarm (PSO) with archive | MOPSO (SilverHawk †) | External repository; DE injection | [35] |
| | TALON (SilverHawk ★) | PSO with Lévy flights + DE injection from the archive | – |
| Portfolio | STRIKE-MO (SilverHawk ★) | Explore-then-commit selection among OSPREY, EAGLE, TALON | – |
| PSO without archive | NSPSO (Opossum) | Non-dominated sorting PSO | [36] |
| Decomposition | MOEA/D (Opossum; SilverHawk †) | Tchebycheff subproblems | [37] |
| Ant colony | MACO (Opossum) | Hypervolume-based ant colony | [38] |
| Model-based | RBFMOpt (Opossum) | RBF surrogate + Tchebycheff decomposition | [30] |

The SilverHawk configurations work on normalised variables, evaluate each generation as one batch, as the Grasshopper execution model requires, and write every non-dominated evaluation to a passive archive used only for reporting. Their deviations from the reference descriptions are listed in Supplementary S2: a smaller SBX distribution index ($\eta_c$ = 2; canonical 15–20) in NSGA-II and NSGA-III, a normalised Tchebycheff function and an external archive in MOEA/D, DE injection in MOPSO, and jDE self-adaptation, Solis–Wets refinement and archive mating in GDE3. The configurations marked ★ combine these components; the supplementary pseudocode gives their structure, and the complete parameter file is available to editors and reviewers on request (Data availability). OSPREY differs from the SilverHawk GDE3 only in the truncation criterion, so that pair isolates hypervolume-contribution selection, and the pymoode GDE3 isolates the effect of the SilverHawk modifications.

**Table 3.** Settings that differ between implementations of the same algorithm, as they apply in the campaign. pymoo: library source, release 0.6.2; pagmo: library defaults of `nsga2` and `moead` [24], which the Opossum 3.1.2 interface does not expose apart from the population size; WallaceiX: defaults of release 2.7 and statements of its developers [39, 40]; SilverHawk: Supplementary S2. n.d. = not documented by the tool. $D$ = 22.

| Setting | NSGA-II pymoo 0.6.2 | NSGA-II Opossum 3.1.2 (pagmo) | NSGA-II WallaceiX 2.7 | NSGA-II SilverHawk 0.6 |
|---|---|---|---|---|
| Crossover | SBX, $p_c$ = 0.9, $\eta_c$ = 15, per-variable probability 0.5 | SBX, $p_c$ = 0.95, $\eta_c$ = 10 | SBX, $p_c$ = 0.9, $\eta_c$ = 20 | SBX, $p_c$ = 0.9, $\eta_c$ = 2 |
| Mutation | polynomial, $\eta_m$ = 20, per variable 1/$D$, applied to 90% of offspring | polynomial, $\eta_m$ = 50, per variable 0.01 | polynomial, $\eta_m$ = 20, per variable 1/$D$; described as bounded and generation-aware | polynomial, $\eta_m$ = 20, per variable 1/$D$ |
| Out-of-bound values | clamped | bounded operators, then clamped | n.d. | reflected |
| Binary variables | continuous, rounded at evaluation | integer sliders; pagmo integer operators if declared as integers (n.d.) | integer sliders | integer sliders, rounded by the slider |
| Duplicate offspring | removed | n.d. | n.d. | kept and counted in the budget; repeats served from a cache in Grasshopper |
| Changed by the user | population size | population size | population size, generations | none |

| Setting | MOEA/D Opossum 3.1.2 (pagmo) | MOEA/D SilverHawk 0.6 |
|---|---|---|
| Weight vectors | simplex lattice (pagmo default), 21 / 36 / 91 | Das–Dennis simplex lattice, 21 / 36 / 91 |
| Variation | DE ($CR$ = 1, $F$ = 0.5) + polynomial mutation, $\eta_m$ = 20 | SBX ($p_c$ = 1.0, $\eta_c$ = 5) + polynomial mutation, $\eta_m$ = 20, 1/$D$ |
| Neighbourhood, mating | $T$ = 20; mating from the neighbourhood with probability 0.9 | $T$ = 20; δ = 0.9 |
| Scalarization | Tchebycheff | Tchebycheff normalised by the |

| Setting | MOEA/D Opossum 3.1.2 (pagmo) | MOEA/D SilverHawk 0.6 |
|---|---|---|
| External archive | none | running ideal and nadir points<br>≤ 200 solutions |

Table 3 sets these implementations side by side. Under the same name, the SBX distribution index spans an order of magnitude (2 to 20) and the per-variable mutation probability a factor of 4.5, so the implementations explore with different strength; bound and integer handling also differ or are not documented, pymoo additionally removes duplicate offspring, and the WallaceiX developers state that its crossover and mutation were adjusted for architectural problems [40]. The two MOEA/D implementations share the weight lattice but not the variation operator: pagmo varies solutions by differential evolution, SilverHawk by SBX. None of these settings was changed in the campaign. Opossum exposes only the population size, and the pymoo defaults are visible only in the library source; WallaceiX shows its crossover and mutation parameters in its interface, but a user who keeps the defaults, as the campaign did, runs a different NSGA-II from the users of the other tools.

**Development history.** The SilverHawk configurations were developed and tested on analytic functions and on an earlier building problem of the same class before this campaign; in particular, their behaviour on the ZDT functions [41] was measured in April–May 2026, five months before the campaign on the present problem, with build 0.6.27, which differs from the campaign builds only in the user interface. The settings were then frozen. Because some adaptation to building problems of this class cannot be excluded, the canonical library algorithms are included as a reference that the author did not configure, and the ZDT results are reported in Sections 5.2 and 5.7. A settings ablation inside the canonical pymoo NSGA-II (Section 6.1) uses the same budget, allocations and seed schedule as the library runs.

### *3.2. Budget, allocation and runs*

All population-based configurations receive 5,000 evaluations, split as population × generations = 20 × 250, 40 × 125 or 100 × 50; 40 × 125 corresponds to earlier architectural benchmarks [19] and the extremes bracket what a practitioner would plausibly choose. MOEA/D runs 21, 36 and 91 subproblems in both tools (Table 3; both use a three-objective simplex lattice, whose sizes these are). All configurations start from random populations (uniform; Latin hypercube in pymoode) and stop only on the budget. RBFMOpt receives 500 evaluations, the budget of the original simulation-based benchmark, because model-based methods converge within a few hundred evaluations on building problems [18, 19] and their computation per iteration grows with the number of evaluated points [17]; one run was extended to 2,000 evaluations to test whether a larger budget changes the comparison. The complete settings are listed in Supplementary Table S2.

The tool configurations were run in Grasshopper for Rhinoceros 8 (SR35) on one workstation (Intel Core i7-4770K, 3.5 GHz, 32 GB RAM, Windows), five times per allocation, the replication of the original study [2]. The SilverHawk configurations were also run ten times per allocation in the headless console with a fixed seed schedule ($5000 + 137 \cdot r$). The library configurations, which are cheap to run outside Grasshopper, were run 30 times per allocation with the same seed schedule; the first five runs are used wherever equal sample sizes are needed, and medians are reported over all 30.

## 4. Indicators and statistics

### *4.1. Indicators*

The primary indicator is the exact hypervolume (HV) [42, 43]. Because HV rankings can depend on the reference point [44], its rule was fixed before the final campaign: $r = 1.1 \times$ the per-objective worst value over all evaluations of the tool configurations, giving $r$ = (58.20 kWh/(m²·a), 2,228.38 K·h/a, 0.0802); the library runs were added later and do not move $r$. HV is computed on two sets per run: **HV-archive**, the non-dominated subset of all evaluations, i.e. what a designer can retrieve from the run; and **HV-final**, the non-dominated subset of the final population (for RBFMOpt, which has no population, the archive). Secondary indicators are IGD+ [45] on normalised objectives against the best-known front, the anytime HV-archive at fixed evaluation counts, and wall-clock time.

HV and IGD+ measure the whole front, whereas a designer usually looks for designs that meet targets. As a design-oriented indicator, each run's archive is queried for the lowest $Q_{tot}$ that satisfies a target on overheating and daylight, for a grid of nine targets ($K_h \leq 100$, 250, 500 K·h/a × DF ≥ 0.10, 0.15, 0.20). A run that contains no design meeting a target counts as missing it. The result is compared with the best value reached by any configuration and with the emulator's RMSE for $Q_{tot}$ against TRNSYS (2.84 kWh/(m²·a), Table 1): a difference smaller than this error cannot be distinguished by the surrogate on which the study rests, and therefore has no reliable design consequence.

### *4.2. Statistics*

The procedures follow the recommendations for comparing stochastic optimizers [23, 46]; all tests are two-sided with $\alpha = 0.05$.

1. **Omnibus test.** Kruskal–Wallis on HV-archive per allocation [47].
2. **Contrasts.** Exact Mann–Whitney $U$ tests [48] with Holm correction [49] and the Vargha–Delaney $\hat{A}_{12}$ effect size [50], for three families fixed in advance: implementations of the same algorithm (NSGA-II, MOEA/D, GDE3); OSPREY vs SilverHawk GDE3 (truncation criterion); and each SilverHawk configuration vs the canonical library algorithm of its family.
3. **Uncertainty.** Percentile bootstrap 95% confidence intervals of the median HV (10,000 resamples) [51].
4. **Pooling and tiers.** A permutation test stratified by allocation ($5 \cdot 10^4$ permutations) compares two configurations over all allocations without mixing them. The configuration with the smallest mean gap to the per-allocation best median is the reference; configurations not separable from it (Holm-corrected) form the *top tier*, and the others are grouped by mean gap (≤ 3%: second tier; above 3%: third tier).
5. **Allocation effect.** HV across the three allocations is compared per configuration with Kruskal–Wallis (Holm over configurations); for two implementations of the same algorithm, the difference in their change from 20 × 250 to 100 × 50 (an implementation × allocation interaction) is given with a percentile bootstrap 95% CI (20,000 resamples per cell).

6. **Robustness.** The ranking of medians is recomputed with DF clipped at zero (the surrogate produces slightly negative DF values), with an alternative reference point at 1.1 × the nadir of the best-known front on normalised objectives [44], and with IGD+ against a best-known front that excludes all SilverHawk solutions; agreement is measured by Kendall's τ.

**Design-space analysis.** Total-effect Sobol indices [52] of each objective with respect to each variable are estimated on the emulator with Saltelli sampling ($N$ = 8,192, 196,608 evaluations; SALib [53]), with insulation thickness restricted to the data-supported range.

The analysis is scripted in Python (NumPy, SciPy, pymoo) and released with the data. Portions of these scripts, and of the plug-in code that implements the evaluated configurations, were drafted and debugged with the help of Claude (Anthropic; Sonnet 4.6 and Opus 4.6, May–June 2026; Claude 5.5, October 2026); every numerical result was checked against the released data and scripts, as stated in the declaration of generative AI use before the references.

## 5. Results

### *5.1. Overall ranking*

The omnibus test rejects equal HV-archive distributions in every allocation (Kruskal–Wallis over the population-based tool configurations and RBFMOpt, $k = 15$, $p < 10^{-8}$). Table 4 and Fig. 3 give the results for all configurations.

**Table 4.** HV-archive (× $10^3$; median per allocation, with the 95% bootstrap CI of the median at 40 × 125), mean of the per-allocation gaps to the best median, IGD+ (median, 40 × 125) and tier. Tool configurations: five runs; library configurations: 30 runs. RBFMOpt has no population: the same five runs apply to every allocation and it is not tiered.

| **Configuration** | **Family** | **20 × 250** | **40 × 125 [95% CI]** | **100 × 50** | **Mean gap (%)** | **IGD+** | **Tier** |
|---|---|---|---|---|---|---|---|
| OSPREY (SH) | DE | 41.04 | 41.99 [41.58–42.73] | 41.60 | −0.41 | 0.016 | top (reference) |
| EAGLE (SH) | DE | 41.55 | 41.53 [40.86–42.08] | 41.33 | −0.59 | 0.013 | top |
| GDE3 (SH) | DE | 40.93 | 41.68 [41.22–41.93] | 41.25 | −1.03 | 0.039 | second |
| TALON (SH) | PSO + archive | 40.90 | 41.16 [41.07–41.55] | 40.91 | −1.74 | 0.044 | second |
| STRIKE-MO (SH) | portfolio | 41.09 | 41.45 [40.42–42.27] | 40.38 | −1.78 | 0.023 | second |
| **GDE3 (pymoode)** | DE | 41.01 | 40.90 [40.79–40.98] | 40.58 | −2.12 | 0.035 | second |
| MOPSO (SH) | PSO + archive | 40.71 | 40.66 [40.08–40.93] | 40.36 | −2.73 | 0.041 | second |

| Configuration | Family | 20 × 250 | 40 × 125 [95% CI] | 100 × 50 | Mean gap (%) | IGD+ | Tier |
|---|---|---|---|---|---|---|---|
| **SMS-EMOA (pymoo)** | GA, HV selection | 40.20 | 40.80 [40.44–41.08] | 40.28 | −3.09 | 0.015 | third |
| **NSGA-II (pymoo)** | GA, NSGA | 39.94 | 40.59 [40.32–40.71] | 40.52 | −3.27 | 0.035 | third |
| NSGA-II (WallaceiX) | GA, NSGA | 39.90 | 39.43 [38.90–40.64] | 38.58 | −5.78 | 0.036 | third |
| NSGA-II (Opossum) | GA, NSGA | 37.53 | 39.25 [38.17–40.84] | 40.22 | −6.52 | 0.033 | third |
| NSGA-II (SH) | GA, NSGA | 39.47 | 38.79 [38.38–39.19] | 38.59 | −6.63 | 0.056 | third |
| NSGA-III (SH) | GA, NSGA | 37.62 | 38.45 [37.43–38.64] | 37.00 | −9.65 | 0.044 | third |
| MOEA/D (SH) | decomposition | 38.72 | 37.29 [33.77–38.38] | 33.61 | −12.41 | 0.057 | third |
| MOEA/D (Opossum) | decomposition | 31.25 | 34.12 [31.68–38.09] | 36.04 | −18.97 | 0.101 | third |
| MACO (Opossum) | ant colony | 33.20 | 33.51 [32.52–33.97] | 34.57 | −19.07 | 0.084 | third |
| NSPSO (Opossum) | PSO, no archive | 31.92 | 33.01 [29.92–36.51] | 32.96 | −21.78 | 0.068 | third |
| RBFMOpt (Opossum), 500 FE | model-based | 37.97 | = | = | −8.98 | – | – |

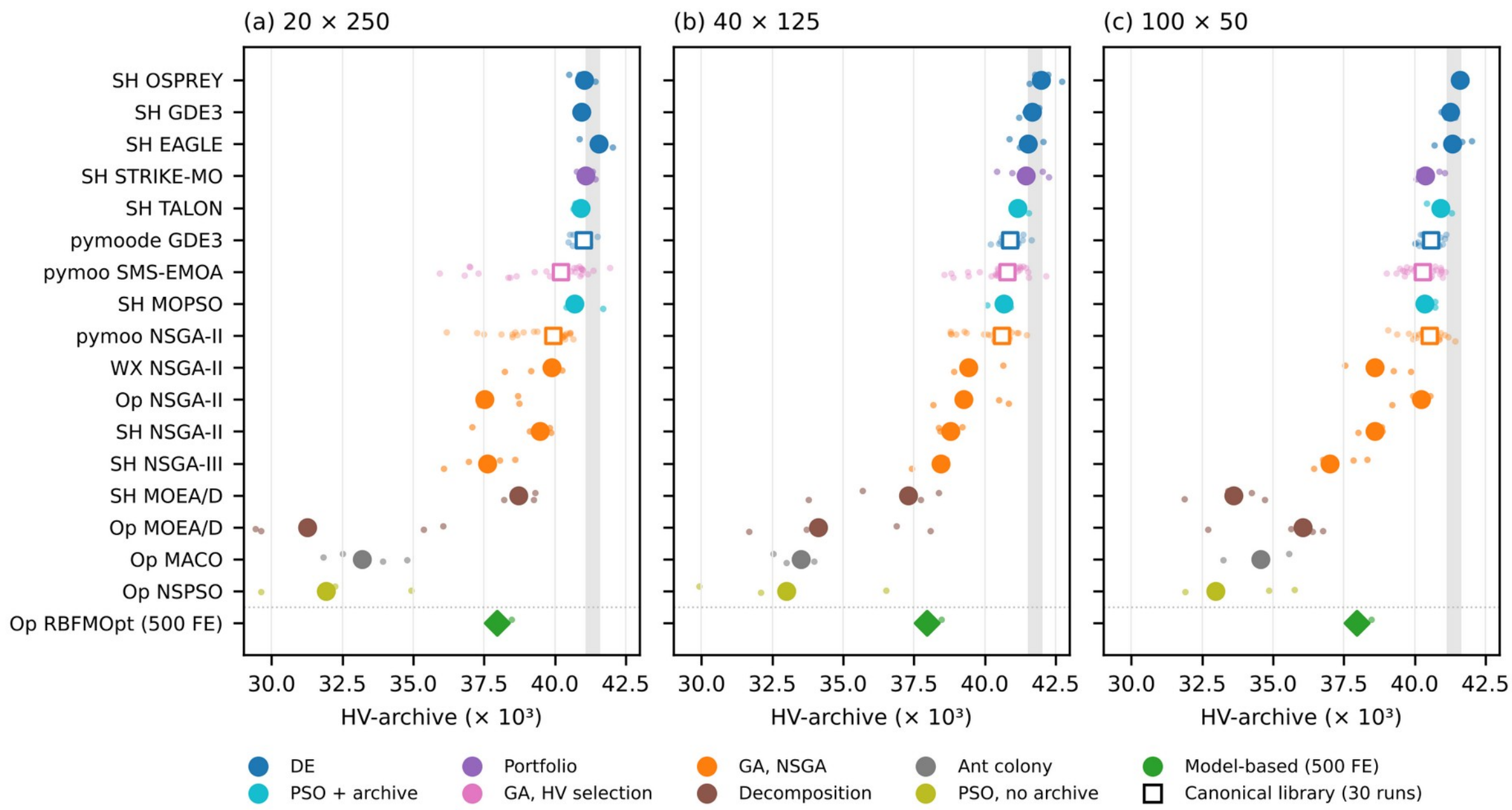

**Fig. 3.** HV-archive of every run by configuration and allocation: (a) 20 × 250, (b) 40 × 125, (c) 100 × 50, ordered by the median at 40 × 125. Filled circles: medians of the tool configurations (five runs); open squares: medians of the canonical library configurations (30 runs); diamond: RBFMOpt at its 500-evaluation budget (the same five runs in every panel); small markers: individual runs; shaded band: within 1% of the best median; colours: algorithm family. SH = SilverHawk, Op = Opossum 3.1.2, WX = WallaceiX.

The top tier is formed by OSPREY and EAGLE, which cannot be separated (stratified permutation test, Holm-adjusted $p$ = 0.70). The second tier, 1.0–2.7% below, contains the other DE, archive-based swarm and portfolio configurations and the canonical GDE3 from pymoode, an implementation the author did not configure. The canonical SMS-EMOA and NSGA-II follow at 3.1% and 3.3%, just beyond the second-tier threshold, and the three NSGA-II implementations of the Grasshopper tools at 5.8–6.6%. The two MOEA/D implementations, MACO and NSPSO lie 12–22% below the best. Configurations that combine DE variation with an archive therefore lead on this problem, and an independent implementation confirms the DE result; but the margin of the leading configurations over the canonical algorithms is about 2–3%, much smaller than the 6–22% that separates them from the weaker configurations of the tools.

### *5.2. Same algorithm, different implementations*

Table 5 contrasts implementations of the same algorithm, whose settings are listed in Table 3, and Fig. 4a shows how each responds to the allocation. The four NSGA-II implementations span up to 6.4% within an allocation; the differences that are significant after Holm correction reach 5.0% (pymoo vs SilverHawk and vs WallaceiX at 100 × 50, $p < 0.001$, $\hat{A}_{12} \geq 0.97$), more than the three canonical algorithms of different families differ from one another at the same allocation (0.7% at 100 × 50, 0.8% at 40 × 125). The canonical pymoo version, with SBX distribution index 15, is never worse than any tool implementation and is 4.6–5.0% ahead of the SilverHawk NSGA-II ($\eta_c = 2$) at 40 × 125 and 100 × 50 ($\hat{A}_{12} \geq 0.95$). The two MOEA/D implementations differ by up to 24%. The SilverHawk GDE3 and the pymoode GDE3 tie at 20 × 250, and at the larger populations the modifications of the SilverHawk version (self-adaptation, local refinement and archive mating) add 1.6–1.9%; hypervolume-contribution truncation (OSPREY vs SilverHawk GDE3) adds a further 0.3–0.9%, significant only at 100 × 50.

**Table 5.** Contrasts on HV-archive: difference of medians, Holm-adjusted exact Mann–Whitney $p$ (within each family and allocation), and $\hat{A}_{12}$ (probability that a run of the first configuration beats a run of the second).

| Family | Contrast | 20 × 250 | 40 × 125 | 100 × 50 |
|---|---|---|---|---|
| NSGA-II × 4 | pymoo vs SilverHawk | +1.2%, 0.57, 0.71 | +4.6%, 0.002, 0.95 | +5.0%, <0.001, 1.00 |
| | pymoo vs WallaceiX | +0.1%, 0.62, 0.58 | +2.9%, 0.34, 0.75 | +5.0%, <0.001, 0.97 |
| | pymoo vs Opossum | +6.4%, 0.070, 0.85 | +3.4%, 0.51, 0.70 | +0.8%, 0.12, 0.77 |
| | SilverHawk vs Opossum | +5.2%, 0.57, 0.80 | −1.2%, 0.51, 0.24 | −4.1%, 0.032, 0.00 |
| | SilverHawk vs WallaceiX | −1.1%, 0.62, 0.28 | −1.6%, 0.16, 0.08 | 0.0%, 0.84, 0.44 |
| | Opossum vs WallaceiX | −5.9%, 0.16, 0.08 | −0.5%, 0.84, 0.44 | +4.2%, 0.095, 0.92 |
| MOEA/D × 2 | SilverHawk vs Opossum | +23.9%, 0.008, 1.00 | +9.3%, 0.31, 0.72 | −6.7%, 0.095, 0.16 |
| GDE3 × 2 | SilverHawk vs pymoode | −0.2%, 0.63, 0.43 | +1.9%, <0.001, 0.97 | +1.6%, <0.001, 0.98 |
| Truncation | OSPREY vs GDE3 (SilverHawk) | +0.3%, 1.00, 0.64 | +0.8%, 0.15, 0.80 | +0.9%, 0.016, 0.96 |

The implementations also respond differently to the allocation (RQ2; Fig. 4a). From 20 × 250 to 100 × 50 the pagmo NSGA-II gains 7.2% and the pymoo NSGA-II 1.5%, while the SilverHawk and WallaceiX implementations lose 2.2% and 3.3%; the pagmo MOEA/D gains 15.3%, while the SilverHawk MOEA/D loses 13.2% (Kruskal–Wallis $p$ = 0.004, Holm-adjusted 0.056), because with the Das–Dennis lattice it runs 21 subproblems for 238 generations at 20 × 250 but 91 for only 54 at 100 × 50. Tested as an interaction, the change from 20 × 250 to 100 × 50 differs by 9.4 and 10.5 percentage points between the pagmo NSGA-II and those of SilverHawk and WallaceiX (bootstrap 95% CIs [2.2, 11.0] and [3.0, 12.4]) and by 28.5 points between the two MOEA/D implementations ([11.1, 38.0]); individually, the gain of the pagmo NSGA-II and the loss of the SilverHawk MOEA/D are significant (exact Mann–Whitney $p$ = 0.008 each), whereas the losses of the SilverHawk and WallaceiX NSGA-II are not. The leading DE and swarm configurations change by less than 2.5% across allocations. Advice such as “use NSGA-II with a population of 100” therefore depends on which NSGA-II is meant. The implementations differ in operator settings that the user does not see; for NSGA-II the SBX distribution index ranges from 2 to 20 and the mutation probability from 0.01 to 0.045 across the four (Supplementary Table S2c); whether these settings explain the differences is tested in Section 6.1.

The divergence is not specific to the building problem. On the ZDT functions run under the same protocol (Section 5.7; ten runs per configuration), the median IGD of the four NSGA-II implementations spans a factor of 3.3–5.5 on ZDT1–3, 11 on ZDT4 and 470 on ZDT6, and 27 of the 30 pairwise contrasts are significant after Holm correction (Supplementary Table S13). The order of the implementations changes with the function: the canonical pymoo NSGA-II is best on ZDT1–3 and third on ZDT6, where WallaceiX is best. The two MOEA/D implementations reverse as well: the pagmo version is 6–35 times better on ZDT1–3, the SilverHawk version 43 times better on ZDT6. The SilverHawk NSGA-II ranks last on every function.

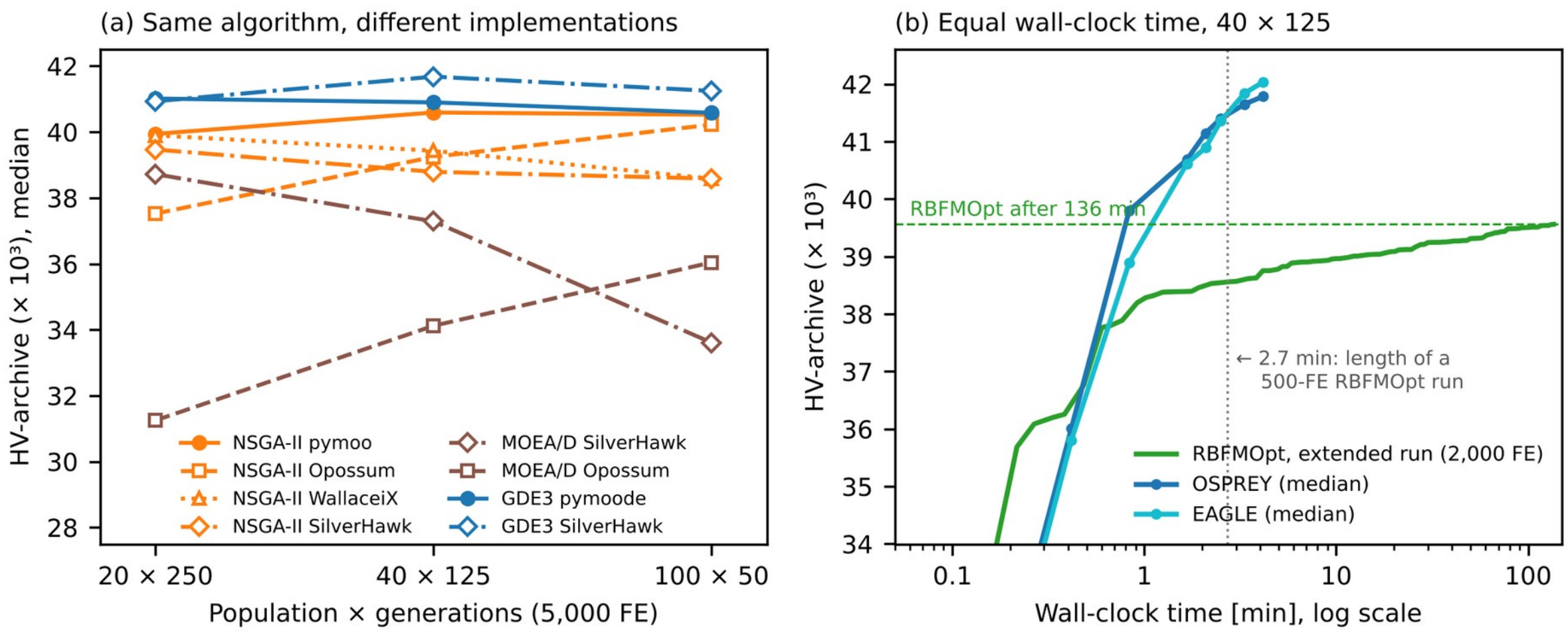

**Fig. 4.** (a) Median HV-archive of the implementations of NSGA-II, MOEA/D and GDE3 at the three population allocations (5,000 evaluations): the same algorithm gains or loses with the population size depending on the implementation. (b) HV-archive at 40 × 125 against wall-clock time, log scale: the RBFMOpt run extended to 2,000 evaluations, from its logged time stamps, and the median console curves of OSPREY and EAGLE placed

on the time axis at the measured canvas cost of 50 ms per evaluation; the dashed line marks the value RBFMOpt reached after 136 min.

### *5.3. Budget in evaluations and in wall-clock time*

At 500 evaluations, the budget of the original simulation-based study, RBFMOpt reaches $37.97 \times 10^3$ at $40 \times 125$, more than every population-based configuration at that point (best: SilverHawk GDE3, $36.81 \times 10^3$; Fig. 5). The DE configurations pass that value within 1,000 evaluations and the NSGA-II implementations after 3,000–4,000. An RBFMOpt run extended to 2,000 evaluations reached $39.56 \times 10^3$; at that budget the medians of OSPREY, EAGLE, GDE3 and TALON were all higher ($40.08–40.69 \times 10^3$). The extended run does not favour the metaheuristics: at 500 evaluations it stood above all five 500-evaluation runs (Supplementary Table S6).

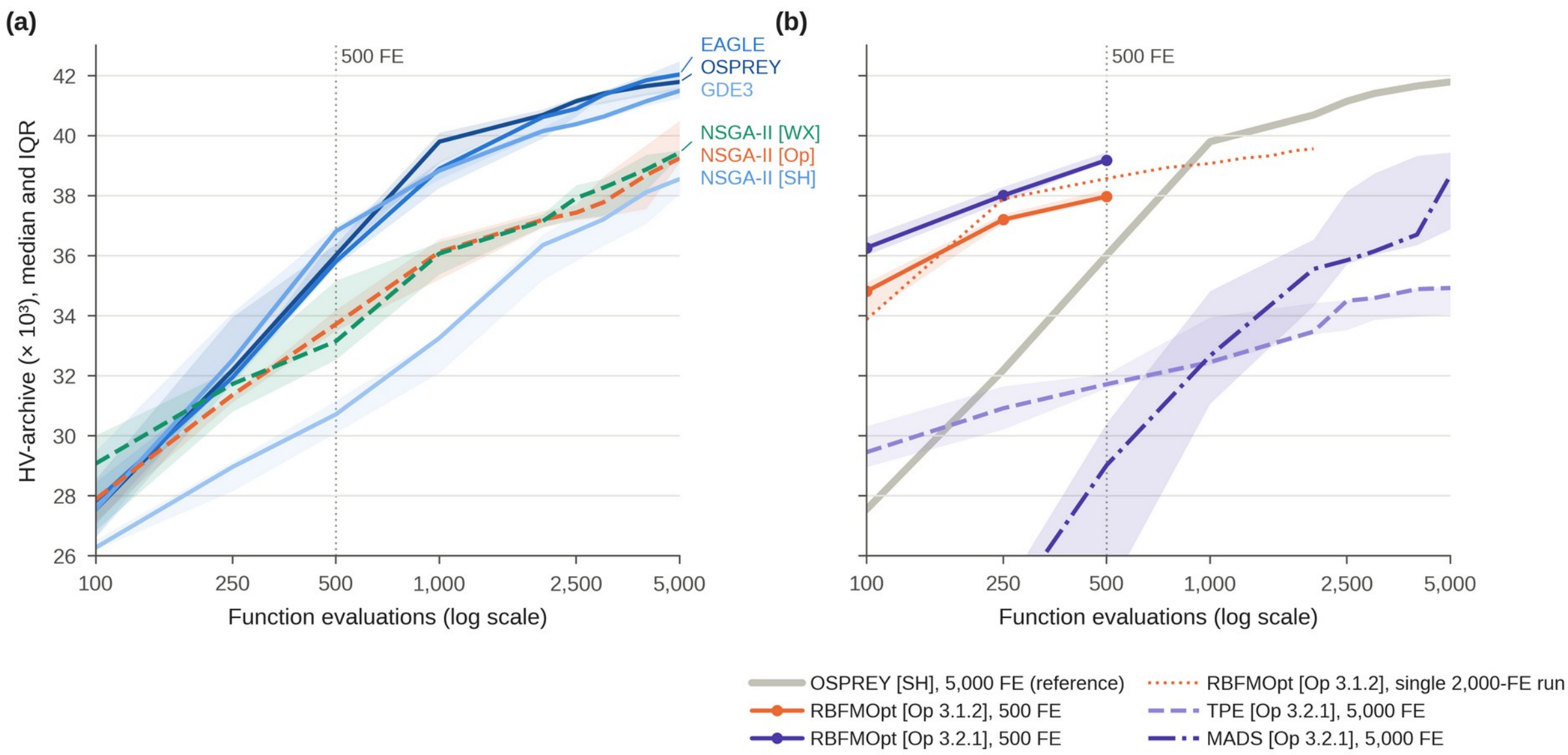

**Fig. 5.** Anytime HV-archive (median and interquartile range) at $40 \times 125$ against the number of evaluations, log scale. (a) Population-based configurations (SilverHawk curves from the console runs, external tools from their canvas logs). (b) Model-based and direct-search methods at their own budgets, with OSPREY at 5,000 evaluations as reference; dotted: single RBFMOpt run extended to 2,000 evaluations.

Counted in wall-clock time, the comparison reverses (RQ3), because the computation per iteration of a model-based method grows with the number of points in its model [17, 30]. The 500-evaluation RBFMOpt runs took 2.7 min on the canvas; the 2,000-evaluation run took 2 h 16 min, as the time per evaluation rose from 0.4 s to 9.4 s. The population-based configurations of SilverHawk and Opossum cost about 50 ms per evaluation throughout (3.6–4.9 min per 5,000-evaluation run; WallaceiX 23–26 min). The decision a designer faces is therefore not 500 evaluations of one method against 500 of another, but what each method delivers in the same time. In the 2.7 min that RBFMOpt needs for 500 evaluations, a population-based configuration completes about 3,200; the DE configurations pass RBFMOpt's 500-evaluation value within 1,000 evaluations, i.e. in less than one minute. Placed on a common time axis (Fig. 4b), the median OSPREY run exceeded after one minute the final value of the 136-minute RBFMOpt run, and at 2.7 min it was 7.6% ahead (Supplementary Table S9). The extended result rests on a single run, and the time axis of the SilverHawk curves combines the console order of evaluations with the measured canvas cost per evaluation (±15% between runs); neither affects the order of magnitude of the difference. The library algorithms are not placed on this axis, because they run outside Grasshopper; their per-evaluation overhead in pymoo is below that of the canvas.

### *5.4. Differences in design terms*

Fig. 6 translates the archives into the quantity a designer asks for: the lowest energy demand reachable under an overheating and a daylight target, summarised as the median gap to the best configuration over a grid of nine targets (values per target in Supplementary Table S10). At 40 × 125, EAGLE, OSPREY and the canonical SMS-EMOA reach the best value of each target to within 0.4 kWh/(m²·a), and eleven further configurations to within 2.0 kWh/(m²·a), including all four NSGA-II implementations and the canonical GDE3; for example, under $K_h$ ≤ 250 K·h/a and DF ≥ 0.15 the median runs reach 14.0 (EAGLE), 14.7 (SMS-EMOA), 15.5 (pymoode GDE3) and 15.8 kWh/(m²·a) (pymoo NSGA-II). These differences are smaller than the emulator's error for $Q_{tot}$ against TRNSYS (RMSE 2.84 kWh/(m²·a)), so within this group the choice of optimizer does not change the design in a way the underlying model can resolve. The configurations that fall outside are the two MOEA/D implementations, MACO and RBFMOpt at 500 evaluations, which are 2.6–3.3 kWh/(m²·a) behind and, together with NSPSO and the SilverHawk NSGA-III, miss the strictest target ($K_h$ ≤ 100 K·h/a with DF ≥ 0.10) in the median run. The same pattern holds at the other allocations (Fig. 6a, c).

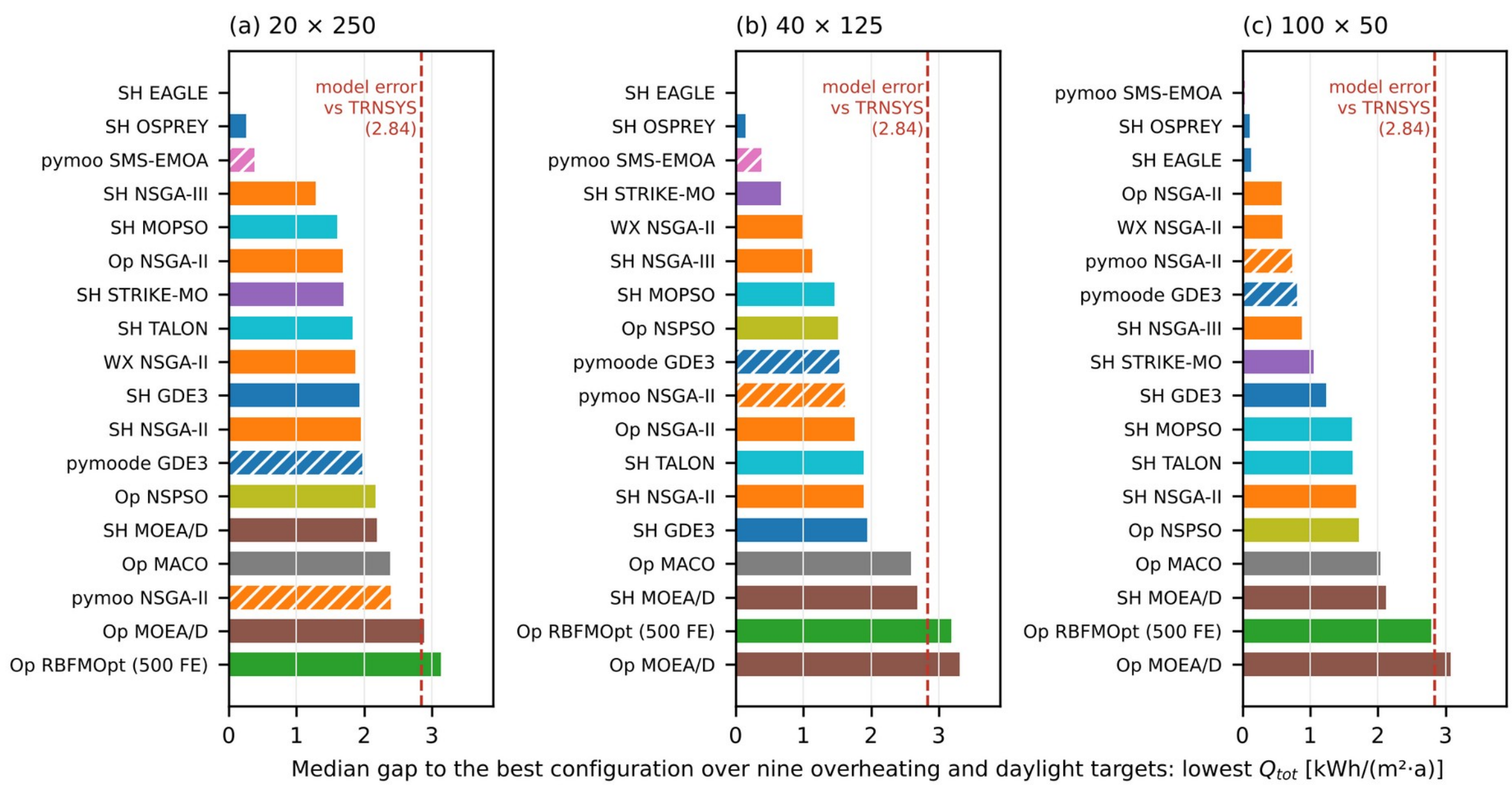

**Fig. 6.** Differences in design terms. For each configuration and allocation, the median over nine targets ($K_h$ ≤ 100, 250, 500 K·h/a × DF ≥ 0.10, 0.15, 0.20) of the gap between the lowest $Q_{tot}$ its runs reach under the target (median over runs; a run that misses the target counts as worst) and the best value reached by any configuration. Dashed line: emulator RMSE for $Q_{tot}$ against TRNSYS. Hatched bars: canonical library configurations; colours: algorithm family as in Fig. 3.

The design-oriented ranking differs from the HV ranking. The canonical SMS-EMOA, eighth by HV, is third here, and the SilverHawk GDE3, third by HV, is fourteenth: SMS-EMOA concentrates its solutions in the low-overheating region where the targets lie, whereas HV also rewards the high-overheating, low-energy end of the front, which the DE configurations cover more completely. Which optimizer is best thus depends on what part of the front the designer needs.

### *5.5. Indicator, reported set and robustness*

Reporting only the final population changes the results substantially (RQ4). From archive to final population the pagmo algorithms lose 2.6–33% of HV, WallaceiX 0.6–4.8% and the library NSGA-II and GDE3 up to 7% and 10%; at 20 × 250 the SilverHawk NSGA-II and GDE3 also lose 12%, while configurations with hypervolume-based selection lose at most 3.5%. The ranking by HV-final agrees only moderately with that by HV-archive (Kendall's $\tau = 0.69$, 0.85 and 0.82). IGD+ agrees with HV at the top and bottom but reorders the middle ($\tau = 0.56$–$0.69$): it favours evenly spread fronts such as those of SMS-EMOA.

The HV ranking is robust to the choices that a sceptical reader might question. Clipping DF at zero leaves the ranking of medians unchanged ($\tau = 1.00$ in every allocation); 6% of the archived points and 6.4% of the best-known front have DF < 0, mostly at the low-overheating extreme reached by the decomposition methods. Moving the reference point to 1.1 × the nadir of the best-known front on normalised objectives gives $\tau = 0.94$–$0.97$. IGD+ computed against a best-known front without any SilverHawk solution ranks the configurations as IGD+ against the full front ($\tau = 0.97$–$1.00$), so the secondary indicator does not depend on the plug-in that supplies most of the front points. Restricting each archive to insulation thickness ≤ 0.44 m, the data-supported range, changes the ranking little ($\tau = 0.85$–$0.96$): the share of solutions beyond 0.44 m is 42% for OSPREY and EAGLE and 64% for RBFMOpt, and OSPREY or EAGLE stays first in every allocation (Supplementary Table S4). The console runs reproduce the canvas ranking of the SilverHawk configurations (Spearman's $\rho = 0.98$, 0.87 and 0.93).

### *5.6. What drives the objectives*

The total-effect Sobol indices (Fig. 7) explain why the problem is easier than its 22 dimensions suggest. $Q_{tot}$ is governed by natural ventilation ($S_T = 0.61$), insulation thickness (0.16) and window sill height (0.16); $K_h$ by the night-time air-change rate (0.58) and the ceiling construction (0.24); DF by the X and Y coordinates of shading rows 2–4 (together 0.74) and the sill height (0.26). The five Z coordinates of the shading rows have no measurable effect on any objective ($S_T \leq 0.002$), and rows 1 and 5 little, so the effective dimension is about 12–15. Each objective is controlled by a different, small group of variables, and the trade-off between them is set mainly by three decisions: whether to use natural ventilation, how far to raise the sill and how dense to make the middle shading rows.

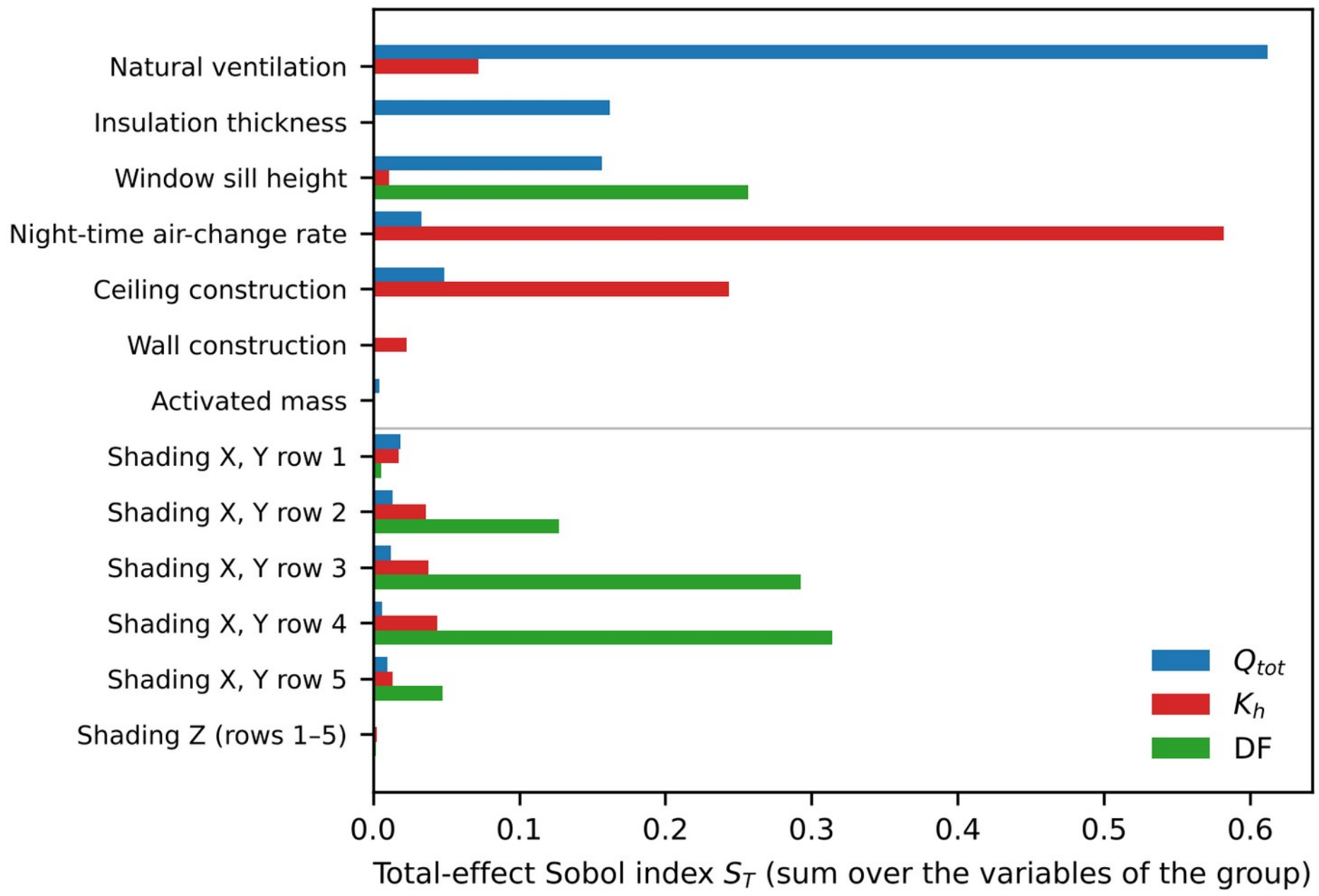

**Fig. 7.** Total-effect Sobol indices of the three objectives on the emulator (Saltelli sampling, $N$ = 8,192), summed over the variables of each group; insulation thickness restricted to the data-supported range [0, 0.44] m.

The fronts reflect this structure. Along the best-known front at 40 × 125, overheating and daylight rise together (Spearman's ρ = 0.86 between $K_h$ and DF), and energy demand falls with the night-time air-change rate (ρ = −0.93) and the insulation thickness (ρ = −0.75); almost all front designs avoid natural ventilation. Designs at the low-overheating end have high sills, thin insulation, low air-change rates and little daylight; designs at the low-energy end have thick insulation, higher air-change rates and more daylight. The configurations populate different parts of this front: all front points contributed by the decomposition methods lie in the lowest quartile of $K_h$, 90% of those of the NSGA family and 44% of those of the leading DE configurations in the highest, whereas SMS-EMOA spreads its points over all four quartiles. The differences between optimizers on this problem are therefore less a matter of finding different kinds of design than of which part of a front, whose shape is set by a few variables, each one populates.

### *5.7. Do rankings transfer from test functions?*

Algorithm choices are often justified by results on analytic test functions. Before the campaign on the building problem, ZDT1–4 and ZDT6 [41] (10 variables, 10,000 evaluations, ten runs) were ported into Grasshopper as black-box components, so that the native solvers of the three tools, RBFMOpt included, ran on them under the same conditions as on the building problem: sliders in, objective values out. The library algorithms were run on the same functions with the same protocol. Fig. 8 compares the ranks (IGD of the final non-dominated set against the true front) with the ranks on the building problem. The weakest configurations are weak everywhere: MACO, NSPSO, both MOEA/D implementations and the SilverHawk NSGA-II and NSGA-III are in the lower half on most ZDT functions and on the building problem. At the top, the order does not transfer (Spearman's ρ between the mean ZDT rank and the building rank = 0.76, RBFMOpt excluded because its building budget differs).

TALON, MOPSO and STRIKE-MO lead on ZDT1–3 and ZDT6, but TALON and MOPSO fail on the multimodal ZDT4, where RBFMOpt is first although it ranks near the bottom on the other four functions; the canonical SMS-EMOA is first on two ZDT functions; OSPREY and EAGLE, the two leaders on the building problem, rank eighth and tenth on average on ZDT. The SilverHawk NSGA-II and NSGA-III rank below the canonical NSGA-II on ZDT as well, so their position on the building problem is not specific to it. The console runs of the SilverHawk configurations on the same functions (30 seeded runs) reproduce the canvas ranks (ρ = 0.88–0.98). The configurations that lead on the building problem were therefore not those that lead on the test functions on which they had been developed and checked.

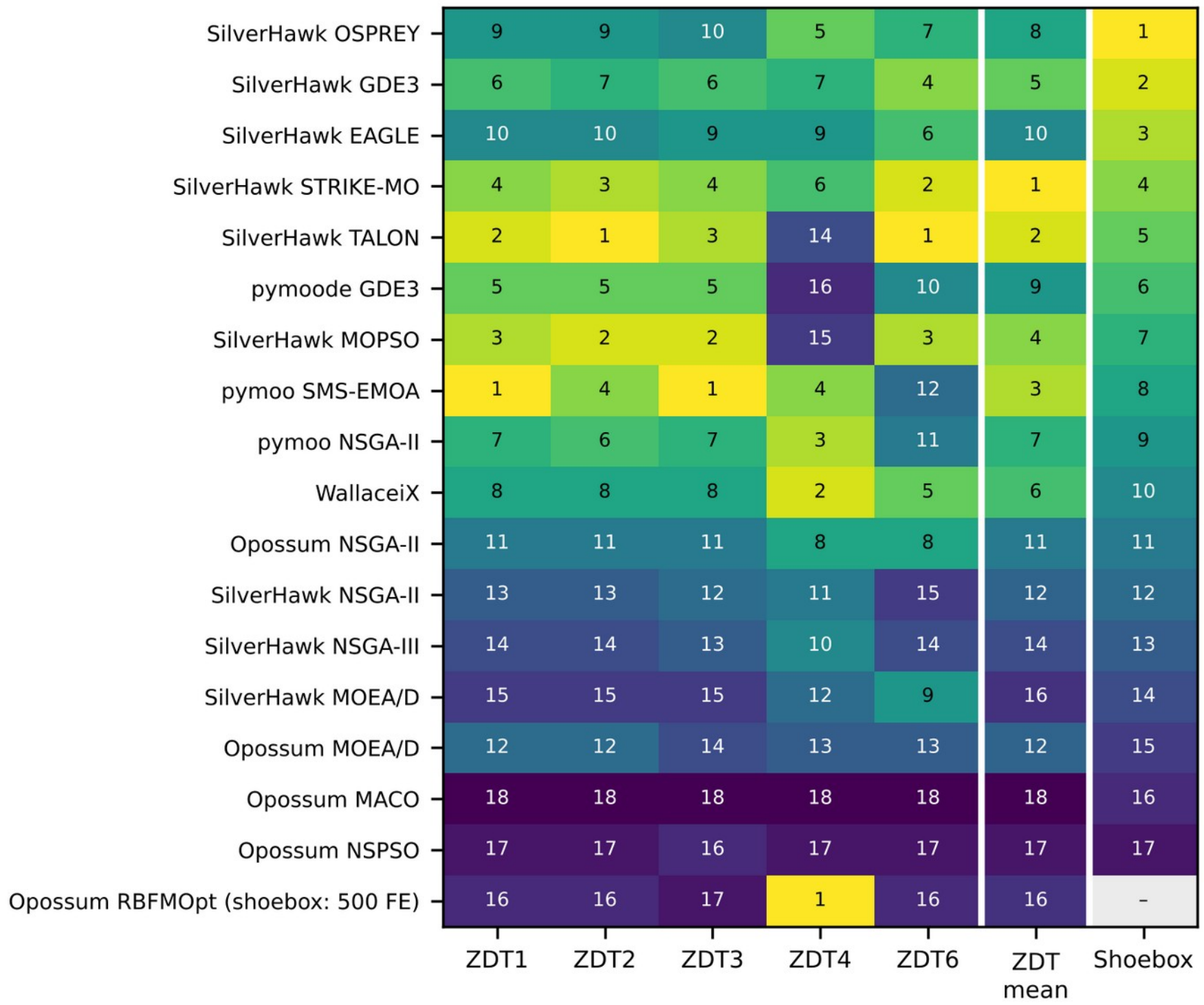

**Fig. 8.** Rank of each configuration (1 = best) on ZDT1–4 and ZDT6 run as Grasshopper black boxes (IGD of the final non-dominated set, $D$ = 10, 10,000 evaluations, median of ten runs), its mean ZDT rank, and its rank on the building problem (HV-archive, 40 × 125). RBFMOpt is not ranked on the building problem, where it ran 500 evaluations. Spearman's ρ between the mean ZDT rank and the building rank = 0.76.

## 6. Discussion

### *6.1. Why implementations of the same algorithm diverge*

The usual question, which algorithm to use, has a less useful answer on this problem than the question of which implementation and which settings. Four implementations of NSGA-II differ by up to 5% in hypervolume after correction for multiple tests, and two implementations of MOEA/D by up to 24%, while three canonical algorithms of different families, NSGA-II, SMS-EMOA and GDE3, lie within 3% of one another (Section 5.2). Configurations that combine DE variation with an archive lead, and a canonical GDE3 that the author did not configure confirms that the DE mechanism, and not only one implementation of it, suits this problem; but the leading configurations gain only about 2–3% over the canonical algorithms, and part of that comes from the hypervolume-contribution truncation, which optimizes the evaluation metric. This is the size of effect that implementation details alone produce within NSGA-II.

Table 3 shows where the implementations part: the strength of exploration set by the SBX and mutation parameters, the handling of bounds, integers and duplicates, and, for MOEA/D, the variation operator itself. To test whether the documented settings explain these differences, they were transferred one at a time, and as complete profiles, into the canonical pymoo NSGA-II and run on the same emulator (30 runs per cell; Supplementary Table S14). The low per-variable mutation probability of pagmo reproduces most of the population response of the Opossum NSGA-II (+5.1% from 20 × 250 to 100 × 50, against +7.2% in the tool; −3.6% at 20 × 250, $p < 0.001$), so for that implementation the mechanism is identified. The documented settings of SilverHawk and WallaceiX do not reproduce their gaps: with $\eta_c = 2$ the pymoo NSGA-II is better, not worse, at 20 × 250 (+1.8%, $p < 10^{-6}$), and the SilverHawk and WallaceiX profiles stay within 0.6% of the library default, whereas both tools lie 4.8% below it at 100 × 50. Their gaps therefore come from details that a settings table does not capture, such as bound handling, selection and the order of evaluation: documented parameters are necessary to identify an implementation, but not sufficient. This also separates the present study from comparisons of optimization frameworks on analytic functions, which traced differences to code-level details by inspection [10]; here the documented settings of closed design tools are tested directly against their measured behaviour.

Two consequences follow, and neither depends on the building problem. For users, a study that reports “NSGA-II” without implementation, version and effective settings cannot be repeated with another tool, and advice on the population size does not transfer between tools. For developers, canonical library algorithms run on a public problem are a cheap regression test: an implementation that falls clearly below them deserves a look at its defaults. Audits of the evolutionary computation literature find the same gap from the other side: settings, seeds and artifacts are among the items most often left unreported [3, 54].

### *6.2. Differences that matter for design*

In hypervolume the configurations span 22%, but in the terms a designer uses the spread is much narrower. For overheating and daylight targets across the range of the problem, fourteen of eighteen configurations reached the best attainable energy demand to within 2 kWh/(m²·a), less than the error of the surrogate model against the simulation (Fig. 6). On such a problem the decision is less about picking the best optimizer than about avoiding the families that fail: on this problem the decomposition and ant-colony methods, which also ranked low on the ZDT functions, and RBFMOpt when it is held to the small budget it was designed for; the archive-free swarm method came close in energy demand but missed the strictest targets. The choice of indicator matters as much: the canonical SMS-EMOA, eighth by hypervolume, is among the three best configurations for the design targets because it places its solutions where the targets lie. A designer who knows which region of the front is of interest should judge optimizers there, or use an indicator restricted to that region.

The sensitivity analysis adds an explanation and a practical lever. Each objective is controlled by a small group of variables, and five of the 22 variables have no effect on any objective (Fig. 7). Removing such variables, or screening them with a few hundred surrogate calls before optimizing, would reduce the problem dimension by a quarter at no cost; with a surrogate the screening takes seconds. The analysis also shows which design decisions set the trade-offs, information that is lost when only an optimized front is reported.

### *6.3. Budget in evaluations and in time*

Zorn et al. found that model-based optimizers win at 500 simulation-based evaluations and that NSGA-II and NSGA-III are the strongest metaheuristics on the surrogate problem at 10,000 evaluations [2]. The present results agree with the first finding: with expensive simulations, RBFMOpt at a few hundred evaluations remains the better choice. They qualify the second: at 5,000 evaluations NSGA-II and NSGA-III trail the DE configurations, although the canonical NSGA-II comes within about 3% of them. This is not a contradiction, because the original study used a different budget, roster and surrogate.

For a design team the relevant budget is often wall-clock time, and when a surrogate answers in milliseconds the overhead of a model-based optimizer dominates: on the workstation used here, a population-based run exceeded within one minute what an extended RBFMOpt run reached in more than two hours. Zorn et al. stress that practitioners can rarely afford more than one run, which makes robustness decisive [1, 2]. With an embedded surrogate and an optimizer whose cost stays proportional to the number of evaluations, the run-to-run variation that dominates the differences within the leading group (Fig. 3) can be handled by repetition instead: five 5,000-evaluation runs take about 20 minutes in Grasshopper and seconds outside it. The evaluation pipeline also counts: in the original study the surrogate was served to Grasshopper through a local Python server, at about 2 s per evaluation with five runs in parallel per workstation [2], whereas the in-process emulator ran at about 50 ms per evaluation, canvas overhead included. Hardware and load differ, so the ratio is indicative only.

The same reasoning applies beyond surrogates: many analyses used in early design, such as linear structural analyses or daylight-factor estimates, return in milliseconds to a fraction of a second. In this regime the overhead of the optimizer is a property of the software as much as of the algorithm, and it should be reported and compared at equal time. On this problem, the configurations that combined an archive with differential-evolution or swarm variation led at equal evaluations and, given the per-evaluation costs measured on the canvas, also at equal time; the NSGA-II of every tool was outperformed by them, and the decomposition and ant-colony methods fell behind. These statements rest on one building problem and on the ZDT functions, and within the leading group the differences are smaller than the surrogate error (Section 6.2).

### *6.4. A reporting checklist for optimizers in design software*

The results translate into eight items that a study using an optimizer embedded in design software should report, and that a tool should make easy to report. Each is backed by a result of this study.

1. **Implementation and version**, not only the algorithm name (Section 5.2).
2. **Effective operator settings**, including those the tool does not expose (Table 3); they are necessary but not sufficient to identify an implementation (Section 6.1), so they do not replace item 1.
3. **The population allocation**, and more than one if conclusions are drawn about it (Section 5.2).
4. **The budget in evaluations and in wall-clock time**, with the hardware (Section 5.3).
5. **Indicators computed on the archive of all evaluations**, with the reference-point rule fixed in advance (Section 5.5).
6. **Differences in design units**, compared with the error of the evaluation model (Section 5.4).
7. **No selection of optimizers on analytic test functions** as a proxy for the design problem (Section 5.7).
8. **Evaluation logs and seeds** released with the study (Data availability).

Items 1, 2 and 8 coincide with items that audits of the evolutionary computation literature find most often missing [54]; items 3–7 are specific to optimization inside design tools.

### *6.5. Limitations*

1. **Surrogate of a surrogate.** The problem is an emulator of the Zorn et al. surrogate problem; its fidelity against TRNSYS is poor for $K_h$ and DF (Table 1), and it inherits slightly negative DF values from the original surrogate. The conclusions concern how optimizer, implementation, budget and reporting choices change what is retrieved from a realistic building response surface, not the physical office; the ranking is robust to clipping DF at zero (Section 5.5).
2. **One building problem.** The results hold for one climate, orientation and zone. The ZDT comparison shows that the weak configurations are weak on very different problems, but the order at the top did not transfer, and it may not transfer to other building problems either.
3. **Runs and defaults.** Five canvas runs per configuration give within-allocation tests little power: an exact two-sided Mann–Whitney test between two tools cannot fall below $p = 0.0079$, and after Holm correction over six contrasts not below 0.048, so a non-significant contrast between two tools indicates low power, not equivalence. Stratified tests, the interaction CIs, the 30 library runs, the console runs and the ten-run ZDT comparison mitigate this. Tuning could change the ranking, especially for MOEA/D and the pagmo algorithms.
4. **Development history and interest.** The SilverHawk configurations were developed on analytic functions and on an earlier building problem of the same class, and they run in a plug-in developed by the author; some adaptation to this problem class cannot be excluded. The canonical library algorithms, the external tools at their defaults, the ZDT comparison and the released data and analysis code are the safeguards; the margins of the leading configurations over the canonical algorithms (Table 5) should be read as upper bounds for what such adaptation could contribute.

5. **Wall-clock comparison.** The extended RBFMOpt result rests on one run, and the library algorithms were not timed inside Grasshopper.
6. **Settings of closed tools.** Opossum and WallaceiX are distributed as compiled plug-ins; Table 3 reports their settings from library defaults, tool releases and developer statements, and marks those that are not documented. The ablation (Section 6.1) transfers only these documented settings into one open-source NSGA-II; it shows what they can explain, not which undocumented detail accounts for the rest.

## 7. Conclusions

On a published 22-variable, three-objective building design problem, eighteen optimizer configurations, including canonical library implementations, were compared on the same emulator of the published surrogate problem at 5,000 evaluations and three population allocations.

1. On this problem the implementation mattered more than the choice among canonical algorithms. Four NSGA-II implementations differ by up to 5% in hypervolume, and their response to the population size differs by up to 10 percentage points, 28 for two MOEA/D implementations; on the ZDT functions the same implementations differ by up to a factor of 470. Three canonical algorithms of different families lie within 3% of one another. The implementations differ in default settings that most of their interfaces do not expose (Table 3), and these settings explain the behaviour of one implementation but not of the other two, so reporting an algorithm without its implementation, version and settings is not enough to reproduce a result.
2. Counted in evaluations, the model-based RBFMOpt leads at 500; counted in wall-clock time on a millisecond surrogate, population-based optimizers overtake it within minutes.
3. The indicator and the reported set change the conclusions. The design-oriented ranking differs from the hypervolume ranking, and final-population reporting understates some configurations by up to a third.
4. Rankings on analytic test functions did not predict which configurations lead on the building problem, and five of the 22 variables have no effect on any objective.
5. Differential evolution with an archive led, also in an independent canonical implementation, but by 2–3% over the canonical algorithms. In design terms, fourteen of eighteen configurations reached the best energy demand attainable under overheating and daylight targets to within 2 kWh/(m²·a), less than the surrogate's error against simulation; decomposition and ant-colony methods did not.

For software and its users this means: name the implementation and its settings, budget in time whenever evaluations are cheap, report the archive, and judge optimizers in the region of the front that matters for the design (Section 6.4). The emulator, the per-run data and the analysis scripts are released, so that other optimizers and tools can be placed in the same frame.

## Declaration of competing interest

The author develops and maintains SilverHawk, the Grasshopper optimization plug-in in which nine of the eighteen configurations are implemented, and the benchmark component used to evaluate the Grasshopper configurations. SilverHawk is not yet publicly released; it will be distributed free of charge, with all algorithms evaluated in this study included in the free distribution, and the author receives no income from it; the author may, however, benefit reputationally and professionally from its visibility. To limit the influence of this interest, the study (i) uses a benchmark problem and data set published independently by other authors; (ii) includes canonical implementations of NSGA-II, SMS-EMOA and GDE3 from independent open-source libraries, run at their library defaults, and two independently developed Grasshopper tools, Opossum and WallaceiX, run at the defaults of their releases; (iii) documents every deviation of the SilverHawk literature algorithms from their reference descriptions and makes the complete parameter settings available to editors and reviewers on request; (iv) reports results on analytic test functions obtained before the campaign; (v) fixed the indicators, the reference-point rule and the statistical procedures before the final campaign; and (vi) releases all per-run data, models and analysis scripts. The author declares no other known competing financial interests or personal relationships that could have appeared to influence the work reported in this paper.

## Funding

This research did not receive any specific grant from funding agencies in the public, commercial, or not-for-profit sectors.

## CRediT authorship contribution statement

**Luciano Ambrosini:** Conceptualization, Methodology, Software, Validation, Formal analysis, Investigation, Data curation, Writing – original draft, Writing – review & editing, Visualization.

## Data availability

The benchmark problem data of Zorn et al. are available at https://doi.org/10.18419/darus-4532. The per-run data of this study (every evaluation of every run, final populations and archives, seeds and timings), the emulator models with the C# and NumPy inferencers, the scripts of the library runs, the ZDT results and the scripts that reproduce all tables and figures are openly available on Zenodo at https://doi.org/10.5281/zenodo.23157761; they include the Grasshopper definition of the campaign and the script that regenerates the library runs with the same seeds. The complete parameter settings of the SilverHawk configurations are held by the author and are available to editors and reviewers on request. The benchmark component and the SilverHawk configurations are part of the SilverHawk plug-in for Grasshopper (0.6 release series), which is not yet publicly released; its release as a free compiled plug-in is planned after publication, and a build is available to editors and reviewers on request. Until then, every configuration except those of SilverHawk can be re-run with the deposited material and the publicly distributed tools, and all reported analyses can be reproduced from the deposited data, except the additional analyses of Sections 5.2 and 6.1 (implementation contrasts on the ZDT functions, interaction intervals and settings ablation), whose scripts and per-run results are available from the author on request and will be added to the Zenodo record on publication. The headless console used for the seeded runs is an in-house development tool of the author and is not distributed; it links the same algorithm code and emulator as the plug-in, and its complete evaluation logs are part of the data release.

## Declaration of generative AI and AI-assisted technologies in the manuscript preparation process

During the preparation of this work the author used Claude (Anthropic; Sonnet 4.6 and Opus 4.6 in May–June 2026, and Claude 5.5 models in October 2026) to assist with programming (drafting, refactoring and debugging portions of the C# plug-in code and of the Python analysis scripts), with checking bibliographic references, and with drafting and language editing of portions of the manuscript text. After using this tool, the author reviewed and edited all content, verified all numerical results against the released data and scripts, and takes full responsibility for the content of the published article.

## Acknowledgements

The author thanks M. Zorn, L. Claus, C. Frenzel and T. Wortmann for publicly releasing the benchmark data set that made this study possible.